\documentclass{article}

 \PassOptionsToPackage{numbers, compress}{natbib}

\usepackage{amsmath,amssymb}

\usepackage{cite}
\usepackage[preprint,nonatbib]{nips_2018}

 \usepackage[nonatbib]{nips_2018}

\usepackage[utf8]{inputenc} 
\usepackage[T1]{fontenc}    
\usepackage{hyperref}       
\usepackage{url}            
\usepackage{booktabs}       
\usepackage{amsfonts}       
\usepackage{nicefrac}       
\usepackage{microtype}      

\usepackage{graphicx}

\newcommand{\fig}{\textbf{Fig.}}
\newcommand{\tab}{\textbf{Table}}

\newcommand{\supp}{\textbf{Supplementary Information}}

\newcommand{\var}{\tilde}

\newcommand{\E}{\mathbb{E}}

\usepackage{url}
\usepackage{multirow}
\usepackage{setspace}
\usepackage[plain]{algorithm}
\usepackage[table]{xcolor}
\definecolor{tableShade}{HTML}{D8DCE1}   
\definecolor{tableShade2}{HTML}{ECF3FE} 
\usepackage{morefloats}
\usepackage{enumitem}
\usepackage{algpseudocode}
\algnewcommand\algorithmicinput{\textbf{Input:}}
\algnewcommand\INPUT{\item[\algorithmicinput]}
\algnewcommand\algorithmicoutput{\textbf{Output:}}
\algnewcommand\OUTPUT{\item[\algorithmicoutput]}

\usepackage{lineno}

\title{A latent topic model for mining heterogenous non-randomly missing electronic health records data}

\author{
  Yue~Li\quad Manolis~Kellis\\
  Computer Science and Artificial Intelligence Laboratory, MIT  
  Cambridge, MA 02191 \\
  \{\texttt{liyue,manoli}\}@mit.edu\\
}

\begin{document}

\maketitle

\begin{abstract}
Electronic health records (EHR) are rich heterogeneous collection of patient health information, whose broad adoption provides great opportunities for systematic health data mining. However, heterogeneous EHR data types and biased ascertainment impose computational challenges. Here, we present mixEHR, an unsupervised generative model integrating collaborative filtering and latent topic models, which jointly models the discrete distributions of data observation bias and actual data using latent disease-topic distributions. We apply mixEHR on 12.8 million phenotypic observations from the MIMIC dataset, and use it to reveal latent disease topics, interpret EHR results, impute missing data, and predict mortality in intensive care units. Using both simulation and real data, we show that mixEHR outperforms previous methods and reveals meaningful multi-disease insights.



\end{abstract}

\section{Introduction}
With recent healthcare initiatives, many hospitals routinely generate large number of electronic health records (EHR) data. In the United States, the number of non-federal acute care hospitals with basic digital systems increased from 9.4 to 96\% over the 7 year period between 2008 and 2015 \cite{Johnson:2016km,Charles2013-wg,Henry2016-ct}. Furthermore, the amount of comprehensive EHR data recording not only clinical notes but also other data types increased from only 1.6\% in 2008 to 40\% in 2015 \cite{Henry2016-ct}. These data types are largely standardized into systematic codes such as Logical Observation Identifiers Names and Codes (LOINC) lab test codes, Diagnosis-Related Groups (DRG) code, International Classification of Diseases (ICD)-9 code, and prescription for electronic drug information exchange code (RxNorm). However, these raw EHR data present several  challenges including interpretation of high-dimensional, heterogeneous and extremely sparse clinical features, and extremely biased measurements based on patient diagnoses and self-reported condition, formally known as non-missing at random (NMAR) \cite{Little2014-nr}.

To address these challenges, we developed a Bayesian unsupervised learning approach that builds on the concepts of collaborative filtering \cite{Salakhutdinov2007-ul,Mnih2008-dk,Hernandez-lobato2014-ju,Marlin2009-ob,Fraser2016-we} and latent topic models \cite{Blei2003-on,Mcauliffe2008-el,Blei2012-vm,Griffiths:2004ey,Teh2007-xx,Asuncion2009-gr}. While it is often infeasible to directly model the joint distribution over the entire EHR feature space, it is possible to formulate a latent topic model over discrete data. Each patient, hereby considered as a ``document", exhibits a mixture of  ``memberships" over a set of latent ``disease topics" each prescribing a distinct spectrum of disease frequency. In this paper, we present MixEHR for modeling heterogeneous EHR data. MixEHR can simultaneously model an arbitrary number of EHR categories with separate discrete distributions while jointly modeling the lab records and the lab results to account for the NMAR observation biases. For efficient Bayesian learning, we develop a variational inference algorithm that scales to large-scale patient EHR data. We apply MixEHR to the largest public dataset called Medical Information Mart for Intensive Care (MIMIC)-III \cite{Johnson:2016km}. We first used MIMIC-III to develop a simulation framework to evaluate MixEHR. We find that the clinical features emerging across EHR categories under common disease topics are biologically meaningful and reveal new insights into disease comorbidity. We also find these disease topics to be highly predictive of mortality.

\section{Related work and our contributions}
Our method is related to the widely popular text-mining method latent Dirichlet allocation (LDA) \cite{Blei2003-on}. However, LDA is not well suited to EHR phenotyping because (1) LDA assumes that the clinical ``terms" are completely observed in the patient ``document", whereas as mentioned above EHR data such as lab tests are often missing; (2) LDA is not suitable to jointly model heterogeneous EHR data categories with distinct distributions; (3) even variations of LDA that deal with missing data do not cope with NMAR mechanism. Our method is also related to several EHR phenotyping studies focusing on matrix factorization and model interpretability \cite{Halpern2016-jc,Joshi2016-hq,Pivovarov2015-xf,Gunasekar2016-gr,Flaherty2005-ht}. These unsupervised approaches differ from the recent supervised deep learning methods that primarily focus on prediction performance of target clinical outcomes \cite{Razavian2015-qd,Cheng2016-vv,Lipton2015-vg,Choi2016-gl,Nguyen2016-fz,Miotto2016-hn,Suresh2017-xx}. Several recent reviews are helpful in gaining perspectives on the recent machine learning advancements in healthcare \cite{Johnson2016-bw,Pathak2013-el,Hripcsak2013-vc}.  \cite{Pivovarov2015-xf} proposed a multi-view LDA with Gibbs sampling for a fixed set of data types.  \cite{Joshi:2016ty} proposed a single-view non-probabilistic modeling of clinical notes, requiring predefined data to learn interpretable disease topics. \cite{Halpern:2016ww} jointly model multiple data matrices but require expert-curated tagging data. \cite{Gunasekar:2016wp} proposed a collective matrix factorization approach on multi-source EHR data. In contrast to these methods, we jointly model lab tests and lab results while inferring missing lab results to account for NMAR bias. Moreover, our joint collapsed variational Bayesian (CVB) inference algorithm provides fast convergence to good local solutions compared to the Gibbs sampling in \cite{Pivovarov2015-xf}. We extend CVB for LDA only \cite{Griffiths:2004ey,Teh2007-xx,Asuncion2009-gr} by inferring the joint expectations of latent topics and lab results not only for the observed but also missing lab tests, which enables us to account for the NMAR observation bias common in the EHR data. We implemented MixEHR in C++ as a standalone Unix command-line software using OpenMP for multi-threaded inference. It allows arbitrary number of discrete data types and discrete states per EHR feature. On a 8-core Ubuntu server, MixEHR takes $\sim$2 hours to learn 100 disease topics from 40,000 patient records over 50,000 EHR features across 6 data types.

\begin{figure}[b]
\begin{minipage}{.4\textwidth}
\includegraphics[width=0.9\textwidth]{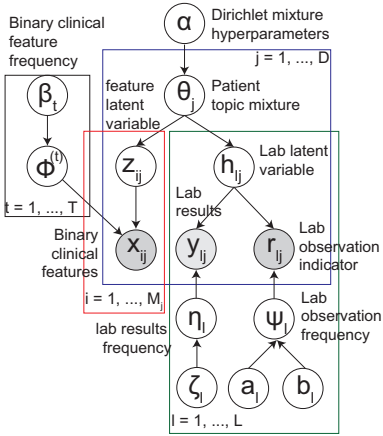}
  \caption{Proposed MixEHR model. See main text for details.}\label{fig:mxr}
  \label{fig:test1}
\end{minipage}%
\begin{minipage}{.5\textwidth}
\includegraphics[width=\textwidth]{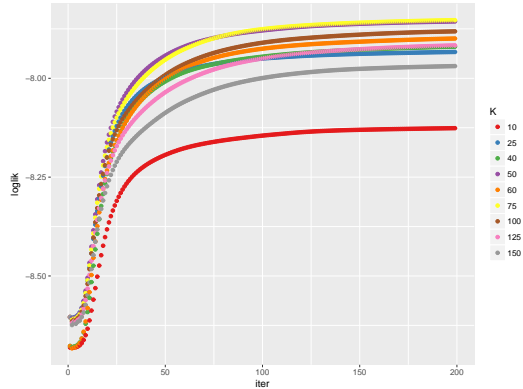}
  \caption{Averaged predictive log likelihood on held-out patients in 5-fold cross-validation.}
  \label{fig:loglik}
\end{minipage}
\end{figure}

\section{MixEHR: a latent topic model for EHR data}
We model EHR data using a generative latent topic model (\fig~\ref{fig:mxr}). Suppose there are $K$ latent disease topics, each topic $k\in\{1,\ldots,K\}$ under data type $t\in\{1,\ldots,T\}$ prescribes a vector of unknown weights $\phi_k^{(t)}=[\phi_{wk}^{(t)}]_{W^{(t)}}$ for $W^{(t)}$ EHR features, which follows a Dirichlet distribution with unknown hyperparameter $\beta_{wt}$. Additionally, each topic is also characterized by a set of unknown Dirichlet-distributed weights $\eta_{lk}=[\eta_{lkv}]_{V_l}$ with $V_l$ distinct values. For each patient $j\in\{1,\ldots,D\}$, the disease mixture membership $\theta_j$ is generated from the $K$-dimensional Dirichlet distribution $Dir(\alpha)$ with unknown asymmetric hyperparameters $\alpha_k$. To generate EHR feature $i$ for patient $j$, a latent topic $z_{ij}^{(t)}$ under data type $t$ is first drawn from multinomial distribution with rate $\theta_j$. Given the latent topic $z_{ij}^{(t)}$, a clinical feature $x_{ij}^{(t)}$ is drawn from multinomial distribution with rate equal to $\phi_{z_{ij}^{(t)}}^{(t)}$. 

For lab data, we use a generative process where each patient has a variable for the result for \emph{every lab test} regardless whether the results is observed. One important assumption we make here is that the lab results $y_{lj}$ and lab observation $r_{lj}$ for patient $j$ are conditionally independent given the latent topic $h_{lj}$, namely that the probability of taking the test and the value of that test both depend on the true disease status, but not on each other. In terms of the generative process, we first sample the latent variable $h_{lj}$ from the multinomial distribution with rate $\theta_j$. Conditioned on the latent variable, we then \emph{concurrently} sample (1) the lab result $y_{lj}$ from $V_l$-dimensional Dirichlet distribution $\eta_{lh_{lj}}$ with hyperparameters $\zeta_l$ over $V_l$ values and (2) the lab observation indicator $r_{lj}$ from Binomial distribution\footnote{Binomial is used such that each lab tests can be performed multiple times on the same patient, which is not uncommon in the real EHR data} with rate $\psi_{lh_{ij}}$, which follows a Beta distribution with unknown shape and scale hyperparameters $a_{l}$ and $b_{l}$. The hyperparameters $\alpha_k, \beta_{wt}, \zeta_{lv}, a_l, b_l$'s follow unknown Gamma distributions.

Formally, we first generate global variables for disease topics:
\begin{align*}
	&\phi^{(t)}_k \sim Dir(\beta_t): \frac{\Gamma{(\sum_w\beta_{wt})}}{{\prod_w\Gamma(\beta_{wt})}}\prod_w[\phi_{wk}^{(t)}]^{\beta_{wt}-1}\\
	&\eta_{lk} \sim Dir(\zeta_l): \frac{\Gamma{(\sum_v\zeta_{lv})}}{\prod_v\Gamma{(\zeta_{lv})}}\prod_v\eta_{lkv}^{\zeta_{lv}-1},\quad
	\psi_{lk} \sim Bet(a_l,b_l): \frac{\Gamma(a_l+b_l)}{\Gamma(a_l)\Gamma(b_l)}\psi_{lk}^{a_l-1}(1-\psi_{lk})^{b_l-1}	
\end{align*}
We then generate local variables for the binary EHR features of each patient:
\begin{align*}
	\theta_j \sim Dir(\alpha): \frac{\Gamma{(\sum_k\alpha_k)}}{\prod_k\Gamma{(\alpha_k)}}\prod_k\theta_{jk}^{\alpha_k-1};
	z_{ij} \sim Mul(\theta_j): \prod_k\theta_{jk}^{[z_{ij}=k]};
	x_{ij} \sim Mul(\phi^{(t)}_k): \prod_w\phi_{kw}^{[x_{ij}=w]}
\end{align*}
and the local variables for the lab data including latent topic, lab result, and observation indicator:
\begin{align*}	
	h_{lj} \sim Mul(\theta_j): \prod_k \theta_{jk}^{[h_{lj}=k]}; 
	y_{lj} \sim Mul(\eta_{lh_{lj}}): \prod_v \eta_{lh_{lj}v}^{y_{ljv}};
	r_{lj} \sim Bin(\psi_{lh_{lj}}): \psi_{lh_{lj}}^{r_{lj}}(1-\psi_{lh_{lj}})^{1-r_{lj}}
\end{align*}	
where $Gam(.)$, $Dir(.)$, $Bet(.)$, $Mul(.)$, and $Bin(.)$ denote Gamma, Dirichlet, Beta, Multinomial, and Binomial distributions, respectively. The notations are in \supp~Table 1.

\section{Variational Bayesian learning of MixEHR model}
\subsection{Marginalized likelihood integrating out $\theta$,$\phi$,$\psi$,$\eta$}
Treating the latent variables as missing data, we can express the complete joint likelihood based on our model as follows:
\begin{align*}
	p(x,y,r,z,h,\theta,\phi,\psi,\eta|\Theta) = 
	p(\theta|\alpha)p(z,h|\theta)p(x|z,\phi)p(\phi|\beta)p(r|h,\psi)p(y|h,\eta)p(\eta|\zeta)p(\psi|a,b)\label{eq:jointlik}
\end{align*}
At this point, we could formulate a variational inference algorithm by optimizing an evidence lower bound of the marginal likelihood with respect to the model parameters \cite{Jordan1998-zj}. However, due to the respective conjugacy of Dirichlet variables $\phi,\eta,\theta$ to the multinomial likelihood variables $x,y,\{z,h\}$, and the conjugacy of Beta variable $\psi$ to the binomial lab observation indicator variable $r$, we can achieve better inference by first integrating out the Dirichlet and Beta variables and then inferring directly the distribution of the latent variables $\{z,h\}$ as follows \cite{Griffiths2004-ma,Teh2007-xx}: 
\begin{align*}
	&p(x,y,r,z,h|\alpha,\beta,\zeta,a,b) =
	\int p(z,h|\theta)p(\theta|\alpha)d\theta
	\int p(x|z,\phi)p(\phi|\beta)d\phi\\
	&\int p(y|h,\eta)p(\eta|\zeta)d\eta
	\int p(r|h,\psi)p(\psi|a,b)d\psi\\
	&= \prod_j\frac{\Gamma(\sum_k\alpha_k)}{\prod_k\Gamma(\alpha_k)}\frac{\prod_k\Gamma(\alpha_k+n^{(.)}_{.jk}+m_{.jk})}{\Gamma(\sum_k\alpha_k+n^{(.)}_{.jk}+m_{.jk})}
	\prod_k\prod_t\frac{\Gamma(\sum_w\beta_{wt})}{\prod_w\Gamma(\beta_{wt})}\frac{\prod_w\Gamma(\beta_t+n^{(t)}_{w.k})}{\Gamma(\sum_w\beta_{wt} + n^{(t)}_{w.k})}\\	
	&\prod_k\prod_l\frac{\Gamma(\sum_v\zeta_{lv})}{\prod_v\Gamma(\zeta_{lv})}\frac{\prod_v\Gamma(\zeta_{lv}+m_{l.kv})}{\Gamma(\sum_v\zeta_{lv} + m_{l.kv})}
	\prod_k\prod_l\frac{\Gamma(a_l+b_l)}{\Gamma(a_l)\Gamma(b_l)}\frac{\Gamma(a_l+p_{lk})\Gamma(b_l+q_{lk})}{\Gamma(a_l+p_{lk}+b_l+q_{lk})}
\end{align*}
where the sufficient statistics are
\begin{align}
	&n_{.jk}^{(.)} = \sum_{t=1}^T\sum_{i=1}^{M_j^{(t)}}[z_{ij}^{(t)}=k],\quad
	n_{w.k}^{(t)} = \sum_{j=1}^D\sum_{i=1}^{M_j^{(t)}}[x_{ij}^{(t)}=w,z_{ij}^{(t)}=k]\\	
	&m_{.jk.} = \sum_{l=1}^L\sum_{v=1}^{V_l}y_{ljv}[h_{lj}=k],\quad
	m_{l.kv} = \sum_{j=1}^Dy_{ljv}[h_{lj}=k]\\
	&p_{lk} = \sum_j[r_{lj}=1]\prod_vy_{ljv}[h_{lj}=k],\quad
	q_{lk} = \sum_j[r_{lj}=0]\sum_v[h_{lj}=k,y_{lj}=v]
\end{align}
Note that we use $y_{ljv}$ to denote the frequency of observing the lab test $l$ of outcome $v$ for patient $j$ and $[y_{lj}=v]$ as binary indicator of a single test. Detailed derivation is in \supp.

\subsection{Joint collapsed variational Bayesian inference}
The marginal likelihood can be approximated by evidence lower bound (ELBO):
\begin{align}
	&\log p(x,y,r|\alpha,\beta,\zeta,a,b) = \log\sum_{z,h} \frac{p(x,y,z,h|\alpha,\beta,\zeta,a,b)}{q(z,h)}q(z,h)\\
	&\ge \sum_{z,h}q(z,h)\log p(x,y,z,h|\alpha,\beta,\zeta,a,b) - \sum_{z,h}q(z,h)\log q(z,h)\\\label{eq:elbo}
	&= \E_{q(z,h)}[\log p(x,y,z,h|\alpha,\beta,\zeta,a,b)] - \E_{q(z,h)}[\log q(z,h)] \equiv \mathcal{L}_{ELBO}
\end{align}
Maximizing $\mathcal{L}_{ELBO}$ is equivalent to minimizing Kullback-Leibler (KL) divergence:
\begin{align*}
	\mathcal{KL}[q(z,h)||p(z,h|x,y,r)] = \E_{q(z,h)}[\log q(z,h)] - \E_{q(z,h)}[\log p(z,h,x,y)] + \log p(x,y,r)
\end{align*}
because $\log p(x,y,r)$ is constant and $\mathcal{KL}[q(z,h)||p(z,h|x,y,r)] + \mathcal{L}_{ELBO} = \log p(x,y,r)$.

Under mean-field factorization, the proposed distribution of latent variables $z$ and $h$ are defined as:
\begin{align}
	\log q(z|\gamma) = \sum_{t,i,j,k}[z_{ij}^{(t)}=k]\log\gamma_{ijk}^{(t)},\quad
	\log q(h|\lambda) = \sum_{l,j,k}[h_{lj}=k]\log\lambda_{ljk}
\end{align}
Maximizing \eqref{eq:elbo} with respect to the variational parameter $\gamma_{ijk}^{(t)}$ and $\lambda_{ljk}$ is equivalent to calculating the expectation of $z_{ijk}^{(t)}$ and $h_{ljk}$ with respect to all of the other latent variables \cite{Bishop:2006ui,Teh2007-xx}:
\begin{align}
	\log \gamma_{ijk}^{(t)} = \E_{q(z^{-(i,j)})}[\log p(x,z)],\quad
	\log \lambda_{ljk} = \E_{q(h^{-(l,j)})}[\log p(y,r,h)]
\end{align}
Normalizing the distribution of $\gamma_{ijk}^{(t)}$ and $\lambda_{ljk}$ gives
\begin{align}
	\gamma_{ijk}^{(t)} = \frac{\exp(\E_{q(z^{-(i,j)})}[\log p(x,z)])}{\sum_k\exp(\E_{q(z^{-(i,j)})}[\log p(x,z)])},\quad
	\lambda_{ljk} = \frac{\exp(\E_{q(h^{-(l,j)})}[\log p(y,r,h)])}{\sum_k\exp(\E_{q(h^{-(l,j)})}[\log p(y,r,h)])}
\end{align}
We can approximate these expectations by first deriving the conditional distribution for $z_{ijk}^{(t)}$ and $h_{ljk}$ (\supp) and then approximating the sufficient statistics by the summation of the variational parameters\cite{Teh2007-xx,Asuncion:2009sm}:
\begin{align}
	\gamma_{ijk}^{(t)} \propto 
	\left(\alpha_{k} + \var{n}_{.jk}^{-(i,j)} + \var{m}_{.jk}\right)\left(\frac{\beta_{tx_{ij}^{(t)}} + [\var{n}^{(t)}_{x_{ij}^{(t)}.k}]^{-(i,j)}}{\sum_w\beta_{wt} + [\var{n}^{(t)}_{w.k}]^{-(i,j)}}\right)\label{eq:gamma}
\end{align}

To infer $h_{lj}=k$, we will need to separately consider whether the lab test $l$ is observed or missing. In particular, we can infer the topic distribution of an observed lab test $y_{lj}$ at value $v$ as:
\begin{align}\label{eq:lambda_obs}
	\lambda_{ljkv} \propto 
	\left(\alpha_{k} + \var{n}_{.jk} + \var{m}_{.jk}^{-(l,j)}\right)\left(\frac{\zeta_{lv}+\var{m}_{lkv}^{-(l,j,v)}}{\sum_{v'}\zeta_{lv'}+\var{m}_{lkv'}^{-(l,j,v)}}\right)\left(\frac{a_l+\var{p}_{lk}^{-(l,j)}}{a_l+\var{p}_{lk}^{-(l,j)}+b_l+\var{q}_{lk}}\right)
\end{align}
For unobserved lab tests, we infer the joint distribution of latent topic and lab result $h_{lj}=k,y_{lj}=v$:
\begin{align}\label{eq:lambda_mis}
	\pi_{ljkv} \propto 
	\left(\alpha_{k} + \var{n}_{.jk} + \var{m}_{.jk}^{-(l,j)}\right)\left(\frac{\zeta_{lv}+\var{m}_{lkv}^{-(l,j,v)}}{\sum_{v'}\zeta_{lv'}+\var{m}_{lkv'}^{-(l,j,v)}}\right)\left(\frac{b_l+\var{q}_{lk}^{-(l,j)}}{a_l+\var{p}_{lk}+b_l+\var{q}_{lk}^{-(l,j)}}\right)
\end{align}
where the notation $n^{-(i,j)}$ indicate the exclusion of variable index $i,j$ and the sufficient statistics are
\begin{align}
	&\var{n}_{.jk}^{-(i,j)} = \sum_{t=1}^T\sum_{i'\ne i}^{M_j^{(t)}}\gamma_{i'jk},\quad
	[\var{n}_{x_{ij}^{(t)}.k}^{(t)}]^{-(i,j)} = \sum_{j'\ne j}^D\sum_{i=1}^{M_{j'}^{(t)}}[x_{ij'}^{(t)}=x_{ij}^{(t)}]\gamma_{ij'k}^{(t)}\\
	&\var{m}_{.jk.}^{-(l,j)} = \sum_{l'\ne l}^L\sum_{v=1}^{V_{l'}}[r_{l'j}=1]y_{l'jv}\lambda_{l'jkv} + [r_{l'j}=0]\pi_{l'jkv}\\
	&\var{m}_{l.kv}^{-(l,j,v)} = \sum_{j=1}^D\sum_{l=1}^L\sum_{v'\ne v}^{V_l}[r_{lj}=1]y_{ljv'}\lambda_{ljkv'} + [r_{lj}=0]\pi_{ljkv'}\\	
	&\var{p}_{lk}^{-(l,j)} = \sum_{j'\ne j}[r_{lj'}=1]\sum_vy_{lj'v}\lambda_{lj'k},\quad
	\var{q}_{lk}^{-(l,j)} = \sum_{j'\ne j}[r_{lj'}=0]\sum_v\pi_{ljkv}
\end{align}
Furthermore, we update the hyperparameters by maximizing the marginal likelihood under the variational expectations via empirical Bayes fixed-point updates \cite{Minka:2000ve,Asuncion:2009sm}). For example, the update for $\beta_{wt}$ is $\beta_{wt}^* \leftarrow \frac{a_\beta - 1 + \beta_{wt}\sum_k\sum_w\Psi(\beta_{wt}+n^{(t)}_{w.k}) - KW_t\Psi(\beta_{wt})}{b_\beta + \sum_k\Psi(W_t\beta_{wt}+\sum_wn_{w.k}^{(t)} - K\Psi(W_t\beta_{wt})}$. Other hyperparameters updates are similar. The learning algorithm therefore follows expectation-maximization: E-step infers $\gamma^{(t)}_{ijk}, \lambda_{ljkv}, \pi_{ljkv}$'s; M-step updates model parameters. To learn MixEHR from massive-scale EHR data in the future, we also developed a stochastic variational inference algorithm \cite{Hoffman:2013tz} (\supp).

\section{Experiments}
\subsection{MIMIC-III dataset}
We applied MixEHR to MIMIC-III data \cite{Johnson:2016km}, which contain $\sim$39,000 patients each with a single admission and $\sim$7500 patients each with multiple admissions (Table 2 in \supp). We used MixEHR to jointly model 6 data categories including unstructured text in clinical notes, ICD-9, DRG, current procedural terminology (CPT), prescriptions, and lab tests, together comprising of $\sim$53,000 clinical features and $\sim$12.8 million total clinical observations. For the 564 lab tests, we used the ``flag" column in the MIMIC LABEVENT data file to record the lab results, which contains empty value indicating normal results or ``abnormal" or ``delta". We set the results of ``delta" values to ``abnormal" as they represent only a small fraction of the patients. As a result, each lab test takes either normal or abnormal values. The lab tests that are not observed for each patient are the missing lab tests. Notably, although we used only two-value system to represent the lab results, MixEHR is able to handle arbitrary number of lab values. We trained and cross-validated our model using the 39k single-admission records and evaluated the trained model using the earlier admission of the 7500 test patients to predict mortality outcomes in their last admissions. 

To evaluate model learning and monitor empirical convergence, we performed 5-fold cross-validation. For each patient in the validation fold, we used randomly selected 50\% of their EHR features to infer their disease mixtures and then used the other 50\% of the features to evaluate the log predictive likelihood, which is a common metric to evaluate topic models \cite{Blei2003-on,Asuncion2009-gr,Hoffman:2013tz}:
\begin{equation}
	\sum_j\sum_{t,i}\log\sum_k\hat\theta_{jk}\hat\phi_{x_{ij}^{(t)}k}^{(t)} 
	+ \sum_l[r_{lj}=1]\log\sum_k\hat\theta_{jk}(\hat\psi_{lk} + \hat\eta_{lky_{lj}})
\end{equation}
where we inferred the variational expectations of the disease mixture and disease topics as:
\begin{align*}
	\hat\theta_{jk} = \frac{\alpha_k + \var{n}_{jk}^{(.)}}{\sum_{k'}\alpha_{k'} + \var{n}_{jk'}^{(.)}},
	\hat\phi_{wk}^{(t)} = \frac{\beta_{wt} + \var{n}_{wk}^{(t)}}{\sum_{w'}\beta_{w't} + \var{n}_{w'k}^{(t)}},
	\hat\psi_{lk} = \frac{a_l + \var{p}_{lk}}{a_l + \var{p}_{lk} + b_l + \var{q}_{lk}},
	\hat\eta_{lkv} = \frac{\zeta_{lv} + \var{m}_{lkv}}{\sum_{v'=1}^{V_l}\zeta_{lv'} + \var{m}_{lkv'}}
\end{align*}
Sensible models should demonstrate improved predictive likelihood on the held-out patients. We evaluated the predictive log likelihood of models with $K \in \{10,25,40,50,60,75,100,125,150\}$ and set \textbf{K=75} as it gave the highest predictive likelihood (\fig~\ref{fig:loglik}).

\begin{figure}
\includegraphics[width=\textwidth]{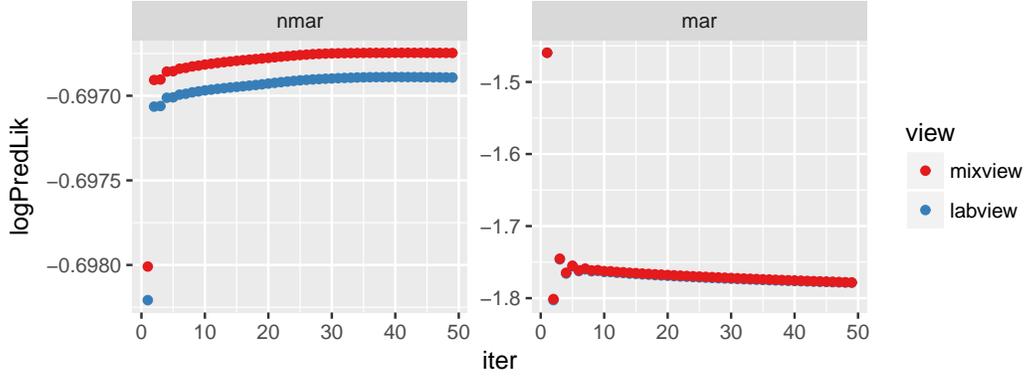}
\caption{Predictive log likelihood of missing lab test results using four different MixEHR models.}\label{fig:sim}
\end{figure}

\subsection{Simulation to evaluate robustness to non-randomly missing lab tests}
To evaluate the novel NMAR aspect of MixEHR, we needed to know the underlying results of the truly missing lab tests for each patient, which is obviously not available. To this end, we simulated EHR data from the presented generative model. Specifically, we ran MixEHR on the full MIMIC data until convergence with the topic number set to 75. We then used the patients' mixtures and the trained model parameters to generate lab data and the other 5 EHR data types. In particular, multinomial latent variable assignments for each EHR variable and each patient were sampled from the given patient mixtures, which were shared across all EHR variables. The missing indicators of each lab test on each patient were simulated from a binomial distribution with the trained beta parameters specific to the latent variable assignments.

We performed 5-fold cross-validation using the simulated data. Because NMAR is only relevant to the lab data, we focused on evaluating the predictive log likelihood on the missing lab test (i.e., $\sum_{l,j}[r_{lj}=1]\log\sum_k\hat\theta_{jk}(\hat\psi_{lk} + \hat\eta_{lky_{lj}}$). Specifically, we used the observed data to infer the mixture of the held-out patients and then used the inferred patient mixtures and the model parameters to impute the missing lab tests. Because the predictive log likelihood is the product of the normalized mixtures and normalized imputed lab results, it is independent from the model complexity and therefore an appropriate evaluation metric. As comparisons, we evaluated four different variations of the model: (1) MixEHR\_nmar modeling NMAR using lab view only (labview); (2) MixEHR\_nmar using all EHR (mixview); (3) \& (4) MixEHR\_mar assuming the lab results are missing at random (MAR) using lab (labview) and all EHR (mixview), respectively. For the MAR model, we did not estimate the distribution of the missing indicator and we only used the observed lab results to update the models as opposed to using both the observed and expected missing lab results.

While the training likelihoods increased in all models, only the NMAR models achieved increasing averaged predictive log likelihood on the missing lab tests for the held-out patients (\fig~\ref{fig:sim}). Moreover, integrating both lab tests and auxiliary EHR data (i.e., clinical notes, ICD-9, prescriptions, etc) provided further improvements. Therefore, our model is robust to NMAR lab results.

\begin{figure}
\includegraphics[width=\textwidth]{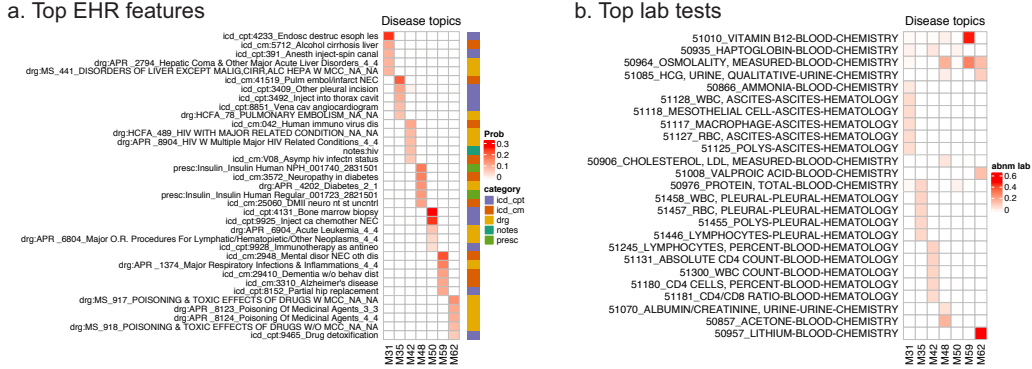}
\caption{Top five EHR features and lab tests for select topics from a 75-topic MixEHR model.}\label{fig:topics}
\end{figure}

\subsection{Multi-view disease topics reveal meaningful pathology}
We examined the clinical relevance of the learned disease topics (i.e., $\phi$, $\psi$, $\eta$) by taking the top 5 EHR binary features and the top 5 lab tests from the same topics. Taking a common index $k$ from those 3 topic matrices gives us one ``disease topic" distribution over the clinical terms across categories. Here we represented the disease topics over lab data using an empirical score calculated as the abnormal lab results weighted by the observation bias normalized over all lab tests for the same topics: $\psi_{lk}\eta_{lkv}/\sum_{l'}\psi_{l'k}\eta_{l'kv}$. Here we sought to intuitively examine how diversified and distinct the disease topics were and how they might help interpreting abnormal lab results. 

For demonstration purpose, we took disease topics that exhibit the highest probabilities over a broad range of diseases including ICD-9 with prefixes \{250, 296, 331, 042, 205, 415, 571\}, which represent diabetes, psychoses, neurodegeneration, HIV, myeloid leukemia, acute pulmonary heart disease, chronic liver disease and cirrhosis, respectively. Notably, these ICD-9 codes are not necessarily the highest ranked features for each topic but we expected prominent enrichments of related clinical terms. Indeed, we observed salient contrast of each topic in terms of the top representative EHR features across diverse EHR categories (colored bars on the right; \fig~\ref{fig:topics}a). Notably, if we were to model the data without modeling distinct distributions over each data category using LDA, the disease topics would have been dominated by clinical notes as they contain overwhelmingly larger number of features (i.e., words) than the other categories. Moreover, we observed interesting connections between abnormal lab results and specific diseases. For instance, topic M31 is enriched for alcohol cirrhosis and blood tests in ascites typical for cirrhosis patients; pulmonary embolism that manifests as blood clotting in lung is linked with proliferation of blood cells via topics M35; under M42, HIV is related with abnormal counts of immune cells such as CD4/CD8 and lymphocytes; diabetes tends to rank high with acetone level under M48, and psychoses topic M62 enriched for opioid abuse is associated with valproic acid and lithium (known markers for psychiatry). Interestingly, topic M59 associated with neurodegenerative diseases such as Alzheimer's disease (AD) is strongly associated with vitamin B12, which was shown to exhibit differential pattern in AD patients \cite{Osimani:JGPN2005}.

\begin{figure}
\includegraphics[width=\textwidth]{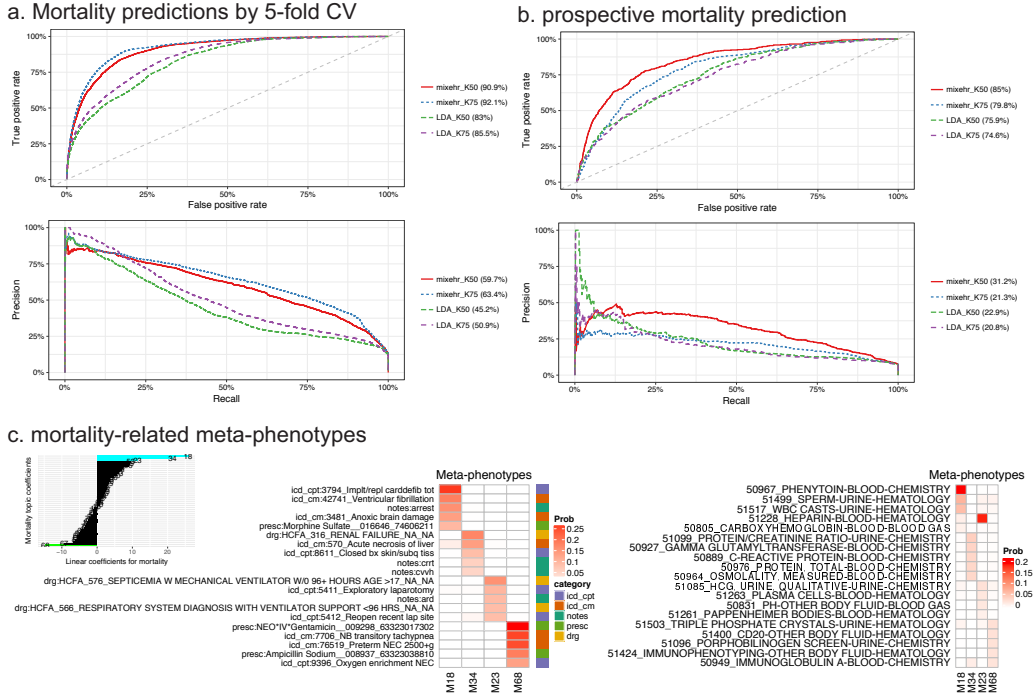}
\caption{Mortality prediction. \textbf{a}. Five-fold cross-validation on predicting mortality of patients with single-admission. \textbf{b}. Predicting patients' mortality outcomes in the last admission using mixture embedding of EHR early admission. \textbf{c}. Topic clinical features of the most and least predictive topics.}\label{fig:death}
\end{figure}

\subsection{Mortality predictions}
Predicting patients' mortality risks in Intensive Care Unit (ICU) is vital to assessing the severity of illness or trauma and facilitating subsequent urgent care. Because we were able to distill meaningful disease topics from heterogeneous EHR data, we sought to assess the predictive power of these disease topics for mortality predictions. For each patient, we labelled them as ``deceased" if their death times were recorded in the admission data. Out of the $\sim$39000 patients with single admission, 4351 (11\%) were deceased. We experimented with $K$=50 and 75 topics. As baseline, we also evaluated LDA implemented in R package ``topicmodels" \cite{JSSv040i13} on flattened EHR matrix of all 6 data types. 

We first performed a 5-fold cross-validation (CV) on patients with single admission by stratifying the patients into five folds with equal proportion of alive and deceased patients. For each CV round, we trained MixEHR on four folds to learn the disease mixture topics. We then applied the trained model to infer the disease mixtures of the training folds, which were then used as predictors to train the supervised LASSO regression on mortality  response variable. Finally, we used the trained LASSO to predict the mortality of the validation fold and repeated the same process for the five folds. Note: (1) mortality labels were not used in learning the disease topics; (2) the unsupervised model were not trained on the test folds. For all CV folds, we recorded the predictions and mortality labels to generate the ROC and precision-recall curves (PRC) and evaluated the area under these curves (AUC). With 75 (50) topics, we obtained 92\% (90.9\%) AUROC and 63\% (59.7\%) AUPRC whereas LDA obtained only 85\% (83\%) AUROC and 51\% (45.2\%) AUPRC (\fig~\ref{fig:death}a; \tab~\ref{tab:death}). 

Moreover, we also performed evaluation on the 7500 patients with multiple admissions. We first trained MixEHR or LDA and LASSO models on the full training set of 39,000 single-admission records. We then inferred the disease mixtures of the test patients from their EHR data in the earliest admission only, predicted mortality outcomes of the patients, and compared the predictions with the death labels in their last admissions. We used only the patients whose last admission was after 30 days but within 6 months of their first admissions, which gave us 458 deceased and 6079 alive patients (7\% positive labels). With 75 (50) topics, we obtained 79.8\% (85\%) AUROC and 21.3\% (31.2\%) AUPRC compared favorably to LDA (\fig~\ref{fig:death}b; \tab~\ref{tab:death}). Based on the LASSO linear coefficients, the top 3 most predictive disease topics associated with mortality are enriched for cardiac arrest, acute liver failure, and septicemia, whereas the least predictive topic is related with  newborn (\fig~\ref{fig:death}c).

\begin{table}[t]
\caption{Mortality prediction}\label{tab:death}
\centering
\begin{tabular}{llrrrr}
\toprule
Data & Method  & \multicolumn{2}{c}{AUROC} & \multicolumn{2}{c}{AUPRC}\\
& & K=50 & K=75 & K=50 & K=75\\
\midrule
5-fold CV & LDA & 83.0 & 85.5 & 45.2 & 50.9\\
5-fold CV & \textbf{MixEHR} & \textbf{90.9} & \textbf{92.1} & \textbf{59.7} & \textbf{63.4}\\
\midrule
Prospective & LDA & 75.9 & 74.6 & 22.9 & 20.8\\
Prospective & \textbf{MixEHR} & \textbf{85} &  \textbf{79.8} &  \textbf{31.2} &  \textbf{21.3}\\
\bottomrule
\end{tabular}
\end{table}

\section{Conclusion and Future Work}
In this paper, we present MixEHR to jointly model heterogeneous EHR data while accounting for NMAR. Our approach demonstrates robustness to NMAR lab results and the ability to distill meaningful disease topics, linking lab results to other EHR features across categories. We also achieved promising mortality predictions. We expect that our method will have broad applications in healthcare. We only demonstrated MixEHR on ICU EHR due to data availability but expect it to also work well on out-patient data. As an extension of MixEHR, we will incorporate longitudinal aspects of the EHR data to model disease progressions (e.g., \cite{NIPS2015_5873,JMLR:v17:15-436}).

\newpage

\bibliography{nips_2018}

\end{document}